%% file: main.tex
\documentclass[10pt,twocolumn,letterpaper]{article}
\usepackage[T1]{fontenc}
\usepackage[pagenumbers]{cvpr}   
\input{cvpr_shipout_compat}
\usepackage{array,tabularx}
\usepackage{cuted}
\definecolor{cvprblue}{rgb}{0.21,0.49,0.74}
\usepackage[hyperfootnotes=false,pagebackref,breaklinks,colorlinks,allcolors=cvprblue]{hyperref}
\input{main_external_labels}
\def\confName{CVPR}
\def\confYear{2027}

\newcommand\blfootnote[1]{%
  \begingroup
  \renewcommand\thefootnote{}\footnotetext{#1}%
  \endgroup
}
\IfFileExists{CJKutf8.sty}{%
  \IfFileExists{u8gbsn.fd}{%
    \usepackage{CJKutf8}%
    \newcommand{\cnorg}{\begin{CJK*}{UTF8}{gbsn}向上跃升（北京）科技有限公司\end{CJK*}}%
  }{\newcommand{\cnorg}{\raisebox{-1.45pt}{\includegraphics[height=8pt]{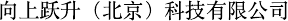}}}}%
}{\newcommand{\cnorg}{\raisebox{-1.45pt}{\includegraphics[height=8pt]{figures/cn_affiliation.pdf}}}}
\hypersetup{pdftitle={FloatGate: Contact-Risk-Aware Visual Updates for Free-Floating Capture in Simulated Spacecraft Cabins},pdfauthor={Jia Deng, Zizhe Wang},pdfsubject={},pdfcreator={}}
\title{FloatGate: Contact-Risk-Aware Visual Updates for\\Free-Floating Capture in Simulated Spacecraft Cabins}
\author{Jia Deng \qquad Zizhe Wang\\
AstroUp Lab, Beijing, China\\
{\tt\small \{dengjia,wangzizhe\}@astroup.com}
}
\begin{document}
\maketitle
\begin{strip}
  \centering
  \includegraphics[width=.92\textwidth]{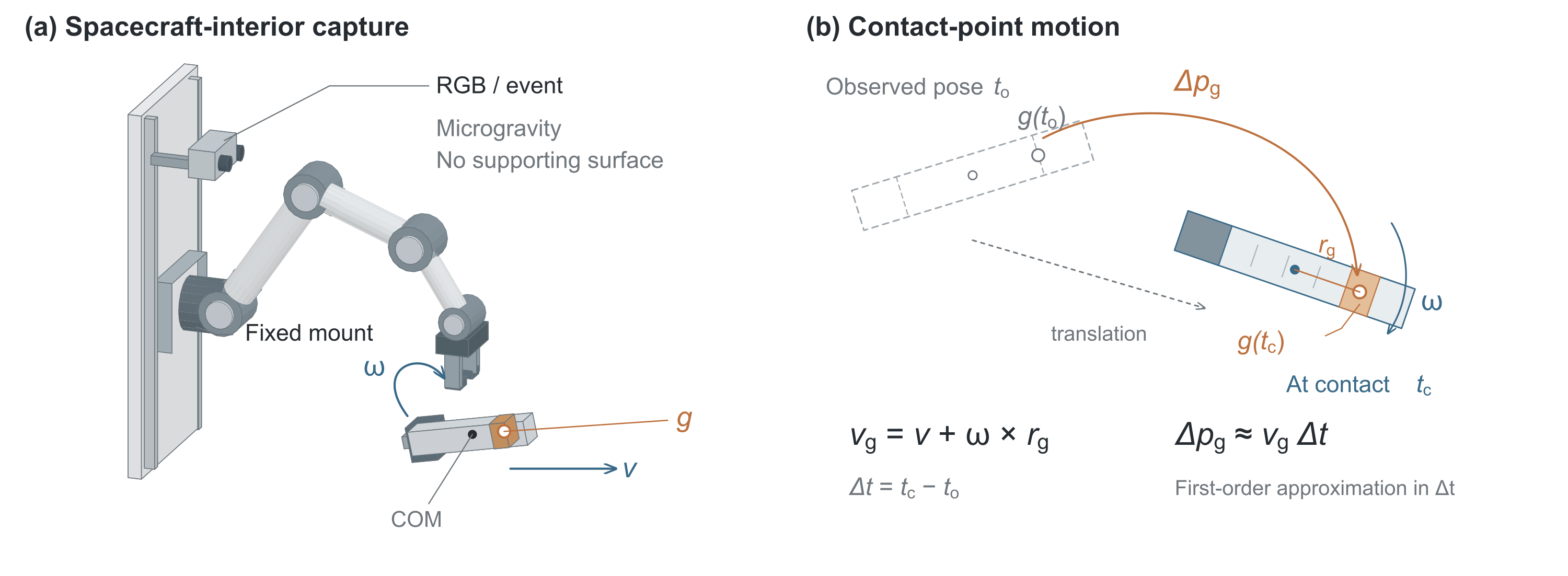}
  \captionsetup{hypcap=false}
  \captionof{figure}{\textbf{Free-floating capture and contact-point motion.} A spacecraft-mounted manipulator observes an unsupported target. Translation and tumbling jointly displace an eccentric grasp reference between observation and contact. The orange connector associates the grasp reference across the two poses.}
  \label{fig:motivation}
  \captionsetup{hypcap=true}
\end{strip}
\input{sections/abstract}
\input{sections/introduction}
\input{sections/related}
\begin{figure*}[t]
  \centering
  \includegraphics[width=.96\textwidth]{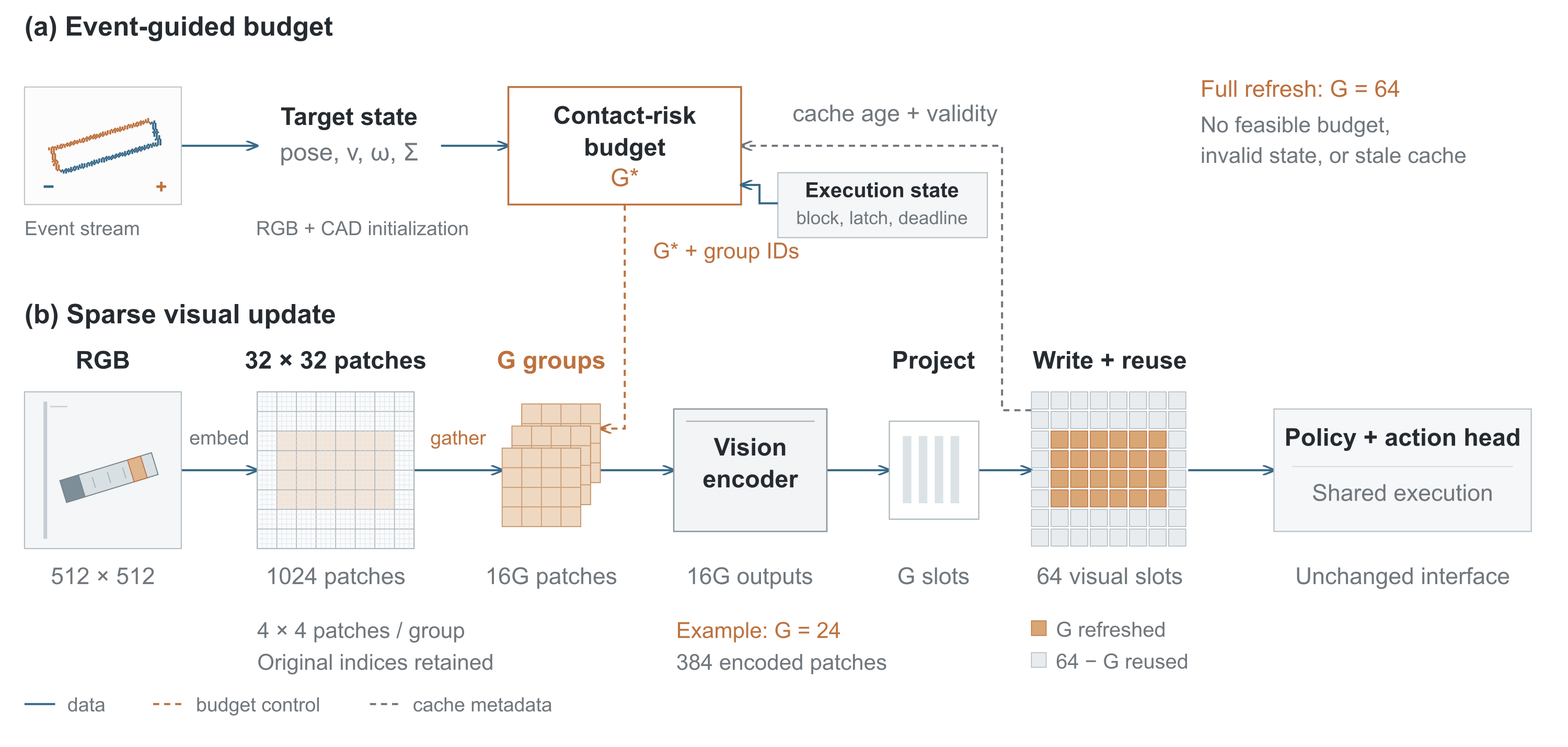}
  \caption{\textbf{FloatGate architecture.} Event-derived state, execution state, and cache statistics determine a contact-error-aware update budget. Complete $4\times4$ patch groups are gathered before visual encoding and scattered into a persistent 64-slot cache. The downstream policy and execution interface are shared across allocation methods.}
  \label{fig:architecture}
\end{figure*}
\input{sections/method}
\begin{figure*}[t]
  \centering
  \includegraphics[width=.96\textwidth]{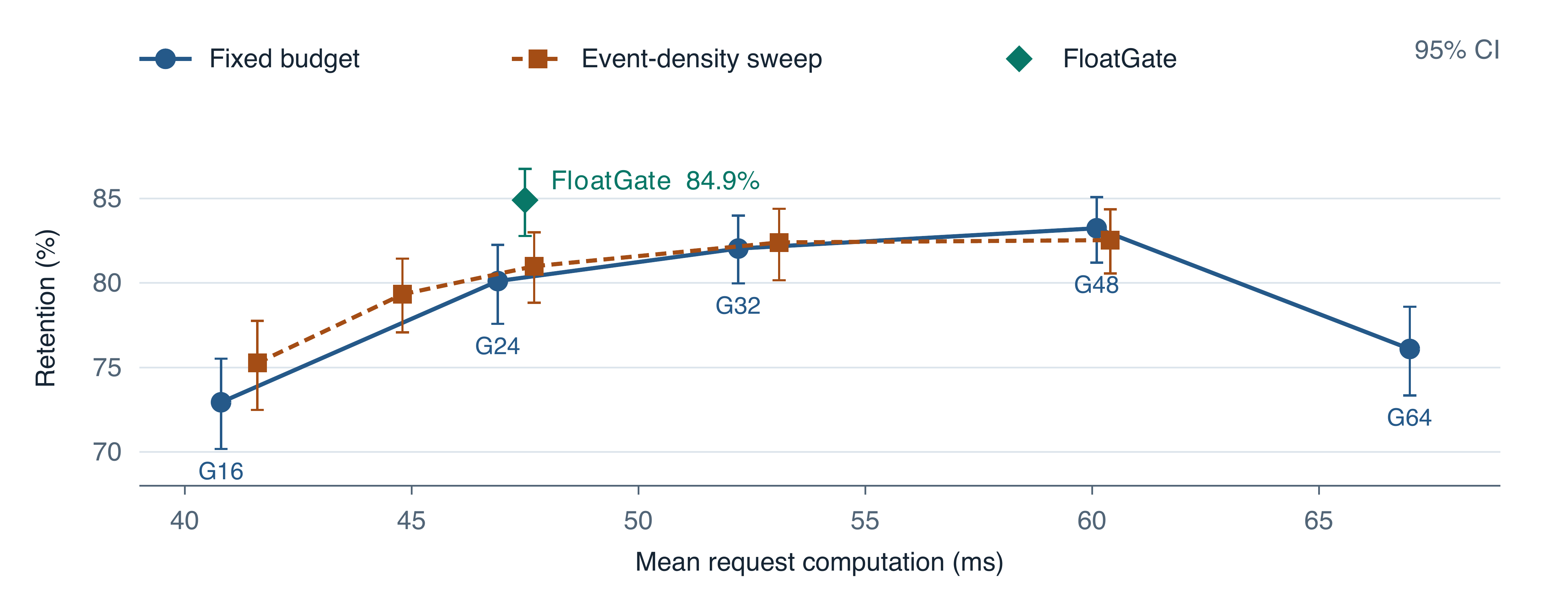}
  \caption{\textbf{Retention--cost trade-off.} Event-fixed sweeps all five budgets; Event-density sweeps validation-frozen thresholds. The contact-risk working point is shown alongside the two sweeps. Error bars are 95\% paired cluster-bootstrap intervals, resampling policy seeds and paired object/start clusters. FloatGate improves retention near the matched-cost operating points.}
  \label{fig:pareto}
\end{figure*}
\input{sections/experiments}
\input{sections/conclusion}
\label{main-text-end}
\clearpage
\input{sections/declaration}

{\small
  \bibliographystyle{ieeenat_fullname}
  \bibliography{references}
}
\end{document}

%% file: cvpr_shipout_compat.tex
\makeatletter
\ifdefined\AddToHook
  \renewcommand*{\@EveryShipout@Init}{}
  \AddToHook{shipout/foreground}{%
    \begingroup
      \unitlength=1pt
      \def\ESO@yoffsetI{0pt}
      
      \ESO@HookIII\ESO@HookI\ESO@HookII
      \global\let\ESO@HookII\@empty
    \endgroup
  }
\fi
\makeatother

%% file: sections/abstract.tex
\begin{abstract}
A free-floating target can translate and tumble while a robot computes a
grasp action.\blfootnote{\cnorg, \href{https://astroup.com}{astroup.com}} Updating more visual information can therefore improve
perception while making the action less timely. We study this trade-off in
simulated spacecraft cabins and introduce FloatGate, a contact-error-based
visual budget selector. Offline candidate branches advance dynamics during
budget-dependent computation and evaluate one committed action block. A
lightweight head conditions on target, cache, and execution state to select
one admissible budget before policy inference. The method reuses sparse
encoding and caching; its contribution is the allocation objective. Across
eight policy seeds and 10,240 paired episodes per method, FloatGate reaches
84.9\% retention versus 76.1\% for full event-assisted updates, while mean
request time falls from 67.0 to 47.5\,ms. A crossed supervision--clock study narrows the gain over execution-aligned
action supervision to 0.89 points (95\% interval $-0.86$--2.64). A separate matched-state audit examines action replacement,
pre-consumption acquisition, and agreement with a local budget oracle.
The point advantage over aligned action supervision and its interaction
with the label clock remain statistically unresolved.
These results concern fixed-base near-field capture with supplied CAD,
not hardware or on-orbit capability.
\end{abstract}

%% file: sections/introduction.tex
\section{Introduction}
\label{sec:introduction}
An unsupported object can move while a robot computes its next action.
Inside a spacecraft cabin, translation and tumbling both displace the grasp
reference; an eccentric contact point can move appreciably even when the
object center barely translates. A delayed or unilateral contact can push
the target out of reach. We study this timing problem in a spacecraft-cabin
simulator with a fixed robot base and known object geometry
(Fig.~\ref{fig:motivation}).

Event observations update motion cues between RGB frames
\cite{gallego2022survey,zhai2026evla}. However, adding these observations does
not by itself shorten a visual policy's computation. Sparse encoding and
caching already offer ways to avoid redundant work
\cite{moon2026actprune,xu2025vlacache,wei2026adaptivecache}. The remaining
question is which update budget best serves a moving contact: refreshing
more information may improve representation while delaying the resulting
action. A central question is whether contact-specific supervision adds value
beyond action losses that are already aligned to execution time.

FloatGate assigns a budget a contact-error target. Offline, five candidate
updates start from the same state and cache. Dynamics continue during each
candidate's profiled computation and execution wait, followed by one
committed action block. Position, normal, and relative-speed errors are
normalized by gripper tolerances. Online, a small head selects the cheapest
admissible candidate before one policy request. Sparse encoding, the
fixed-slot cache, event tracking, and the policy backbone are inherited;
the increment is the contact objective and its budget-dependent time model.

The study tests this increment with shared-interface analytic and learned
allocators, a native latency-aware policy transfer, and eight policy seeds.
A crossed supervision--clock study compares contact and executable-action
losses under identical timing contracts. A separate single-state-per-episode
audit checks replacement, terminal outcomes, and budget decisions; paused
worlds and timing stress delimit the resulting benefit.
The conclusions concern near-field capture with available CAD templates.
They do not establish unknown-geometry manipulation or on-orbit deployment.

%% file: sections/related.tex
\section{Related Work}
\label{sec:related}

\paragraph{Event sensing and dynamic capture.}
Event cameras report per-pixel brightness changes asynchronously, supporting
motion estimation between image observations~\cite{lichtsteiner2008dvs,gallego2022survey}.
E-VLA studies event augmentation for manipulation under dark and blurred
RGB observations~\cite{zhai2026evla}. EDFT uses a known object model and
event-based distance fields for six-degree-of-freedom
tracking~\cite{kang2026edft}. These are reusable perception components for
our task. ESIM couples adaptive rendering with event generation to simulate
contrast-threshold observations~\cite{rebecq2018esim}. Frame-to-event conversion
is useful for visualization, but its temporal support depends on source
sampling~\cite{hu2021v2e}. AstroBench introduces
spacecraft-interior tasks with a fixed-base arm and independently moving
targets~\cite{wang2026astrobench}. We adopt this task setting and distinguish
retention after capture from the benchmark's acquisition metric. Our cabin
setting excludes the coupled base dynamics of free-flying orbital
servicing~\cite{flores2014space}.

\paragraph{Sparse computation and temporal reuse.}
OpenVLA establishes an open, pretrained vision-language-action policy that
can be adapted to robotic manipulation~\cite{kim2024openvla}.
SmolVLA provides a compact policy and an asynchronous inference stack that
separates action prediction from execution~\cite{shukor2025smolvla}.
ActPrune uses the previous action chunk to form geometric priors for
selection before the vision encoder~\cite{moon2026actprune}. FastVLA uses
task-to-vision attention and the uniformity of its importance distribution
to adapt compression across time~\cite{xu2026fastvla}. VLA-Cache reuses
key-value representations of minimally changed visual tokens, with
task-relevant recomputation and layer-dependent reuse~\cite{xu2025vlacache}.
LAC learns a token selector and cache ratio from the VLA task loss,
already providing task-driven allocation~\cite{wei2026adaptivecache}.
EventPrune uses event motion cues for token pruning in dynamic spatial
reasoning~\cite{ma2026eventprune}; our target instead concerns contact
consequences under computation delay. These mechanisms use different interfaces. We compare explicit
common-backbone adaptations, rather than reproductions of their original
pipelines. FloatGate adds a contact-tolerance objective with budget-dependent
physical time. The action-loss control tests this supervision increment
without establishing superiority over LAC's complete optimization.

\paragraph{Latency-aware manipulation.}
Visual servoing provides image- and pose-based feedback
formulations~\cite{chaumette2006visual}. DynamicVLA combines efficient inference with action streaming to address
misalignment between observation and execution~\cite{xie2026dynamicvla}.
VLALagBench studies latency sensitivity across moving targets, policies,
and execution schedules~\cite{wen2026vlalagbench}. These latency-focused studies motivate our distinction between request
completion and control-consumed information age. FloatGate compares
allocation objectives at matched cost. A separate DynamicVLA transfer
retains its native encoder, action horizon, and streaming schedule; this
system comparison does not isolate the budget head.

%% file: sections/method.tex
\section{Contact-Risk-Aware Visual Updates}
\label{sec:method}
FloatGate changes the objective used to allocate visual computation; it does
not introduce a new encoder, cache, tracker, or action model. Its score
estimates the contact error of an action block when the target keeps moving
through that block's computation delay (Fig.~\ref{fig:architecture}).

\subsection{Task and information age}
\label{sec:task}
We study a simulated microgravity cabin with a fixed-base manipulator and an
unsupported rigid target. Known CAD and a fixed object-frame grasp reference
$\mathbf r_0$ are available to every method. Initial states admit a prospective
interception within 500\,ms; the task evaluates near-field capture, not
long-range approach or unknown-shape grasp planning. For target position,
orientation, and twist $(\mathbf p_t,\mathbf R_t,\mathbf v_t,\vec\omega_t)$,
\begin{align}
 \mathbf g_t &=\mathbf p_t+\mathbf R_t\mathbf r_0,\\
 \dot{\mathbf g}_t &=\mathbf v_t+
 \vec\omega_t\times(\mathbf R_t\mathbf r_0).
 \label{eq:graspvelocity}
\end{align}
Thus an eccentric grasp point can move substantially even when its center
moves little. RGB-to-CAD initialization, an EDFT-style event
tracker~\cite{kang2026edft}, and an $\mathrm{SE}(3)$ filter provide estimated
pose, twist, covariance, and innovations. True pose is used only for offline
labels and evaluation.

For request, readiness, and first-consumption times $t_r,t_b,t_u$,
\begin{align}
 T_{\mathrm{req}}&=t_b-t_r,\\
 A_{\mathrm{RGB}}&=t_u-t_{\mathrm{exp}}
 =(t_r-t_{\mathrm{exp}})+T_{\mathrm{req}}+(t_u-t_b).
 \label{eq:age}
\end{align}
Here $t_{\mathrm{exp}}$ is the refreshed frame's exposure midpoint; older
cached features retain separate source times. Request latency includes
selection, transfer, encoding, action generation, and cache maintenance.
An added release wait affects $t_u$ but is excluded from computation.

Under the shared 100\,ms request schedule, shorter computation advances
action replacement within a cycle; it does not increase request frequency.
The executor quantizes availability to 10\,ms ticks, and new observations
cannot enter a request after its snapshot.

A grasp point moving at 0.30\,m/s travels 6\,mm in 20\,ms, comparable to
the 8\,mm alignment tolerance. This scale calculation motivates the target;
it does not predict success, which also depends on estimation and contact.

\subsection{Shared candidates and cache}
\label{sec:interface}
\label{sec:selection}
The SmolVLA-compatible policy~\cite{shukor2025smolvla} receives a
$512\times512$ image with $32\times32$ patches. Complete $4\times4$ patch
blocks form 64 groups, with candidate budgets
\begin{equation}
 \mathcal K=\{16,24,32,48,64\},\qquad G\in\mathcal K.
\end{equation}
A mandatory set $P_t$ covers the grasp reference, target contours, and
gripper sweep. Remaining groups use the common normalized ranking
\begin{equation}
 s_j=0.5c_j+0.25a_j+0.25u_j,
\end{equation}
where $c_j,a_j,u_j$ denote event coverage, cache age, and innovation.
$S_G$ contains $P_t$ and the highest-ranked remaining groups; a budget below
$|P_t|$ is inadmissible. Ranking is shared by the allocation-objective controls.

Gathering $16G$ embeddings before the vision Transformer reduces encoder
work while preserving original positional indices. The connector restores
within-group order and writes each refreshed group to its original slot:
\begin{equation}
 C_{t,j}=\begin{cases}P_j(F_t),&j\in S_G,\\
 C_{t^-,j},&j\notin S_G.
 \end{cases}
 \label{eq:cache}
\end{equation}
Unselected slots retain their values and timestamps. A 250\,ms expiry gate
applies to the target/gripper region and its predicted sweep; inactive
background slots may be older. The downstream policy always consumes 64
ordered slots, so its action-head cost is unchanged. Shared adaptation covers
all budgets and cache histories. Temporal reuse and sparse encoding are
inherited mechanisms, rather than claimed contributions.

Gathering changes attention context despite preserving coordinates.
Adaptation unrolls mixed-budget histories with 20\% full updates,
occlusion, reuse, and full-update feature distillation. RGB-only and
event-conditioned inputs alternate. The active region includes the
protected set and projected target/gripper 500\,ms sweeps. A background slot
entering this region immediately faces expiry; metadata never advances
without recomputation.

\subsection{One-block labels with budget-dependent time}
\label{sec:risklabels}
The executor consumes commands at 100\,Hz, while policy requests occur every
100\,ms. Each policy output contains 50 world-frame Cartesian poses and
gripper commands at 10\,ms intervals, timestamped relative to $t_r$.
While computation runs, the robot follows its preceding block, or holds if
that block ends. At consumption, expired waypoints are removed. A gripper
command crossing 0.5 starts a 50\,ms closure; a persistent latch preserves its
deadline when a later block replaces the Cartesian commands.

\paragraph{Offline branch label.}
Clone robot/target state, estimator, cache, active block, closure latch, and
action randomness into the five candidates. For each admissible budget,
compute its visual update and action block, advance both bodies under the
preceding commands through its profiled duration, and release at the next
executor tick. Commit the remaining candidate block without another policy
request. Its valid closure deadline is $\tau_G$; an expired deadline,
unreachable command, or closure beyond $t_r+500$\,ms is infeasible.
Acquisition by the preceding block before consumption instead cancels the
candidate and enters holding: it is a separate capture-stage terminal
outcome, not an infeasible label. The subsequent retention test still
applies. Contacts always affect dynamics.

At $\tau_G$, compare the executed end-effector with the true moving grasp
reference. Position, oriented-normal, and relative-speed errors define
\begin{equation}
 Y_G=\max\left\{\frac{e_{p,G}}{d_p},
 \frac{e_{n,G}}{d_n},\frac{e_{v,G}}{d_v}\right\},
 \label{eq:label}
\end{equation}
with $d_p=8$\,mm, $d_n=10^\circ$, and $d_v=0.08$\,m/s. The 8\,mm value is a
local alignment tolerance; the separate 10\,mm retention criterion measures
post-acquisition slip over two seconds. Infeasibility has its own binary
label. This target concerns one committed block: normal online replanning
can replace it, so task retention must be evaluated separately.

\subsection{Budget selection and controlled comparisons}
\label{sec:riskhead}
\label{sec:controls}
Candidate features contain estimated motion/covariance, grasp offset,
visibility, event/cache statistics, protected-set size, reaction time,
$G$, and cost $\widehat T(G)$. They also contain measured joint/effector
state, the preceding block's execution summary and remaining waypoints,
and the closure latch/deadline. All allocators receive this execution
state. A shared two-layer, 32-unit MLP predicts a contact-error quantile
$q_\theta$ and infeasibility score $b_\theta$. Feasible labels use
\begin{equation}
 \mathcal L_q=\rho_{0.95}(Y_G-q_\theta(\vec\phi_{t,G})),
 \quad\rho_q(u)=u(q-\mathbf{1}[u<0]).
\end{equation}
Disjoint calibration trajectories supply the nonnegative empirical 0.95
quantile $\delta_G$ of feasible residuals, giving
\begin{equation}
 R_t(G)=\max\{0,q_\theta(\vec\phi_{t,G})+\delta_G\}.
 \label{eq:calibration}
\end{equation}
Residual calibration is related to conformalized quantile
regression~\cite{romano2019cqr}; our trajectory-weighted correction provides
empirical diagnostics, with no finite-sample or post-selection coverage
guarantee.
With state/cache validity $V_t$, batch scoring selects
\begin{align}
 \mathcal F_t=\{G\in\mathcal K:\;&G\geq|P_t|,\ R_t(G)\leq1,\nonumber\\[-2pt]
 &b_\theta(\vec\phi_{t,G})\leq0.1,\ V_t=1\},\\
 G_t^*&=\operatorname*{arg\,min}_{G\in\mathcal F_t}\widehat T(G).
 \label{eq:choice}
\end{align}
An empty set triggers a full refresh and the common hold/recovery rules;
only one selected policy request runs online. Source timestamps are committed
only with refreshed features.

\subsection{Allocation objectives and fitting}
\label{sec:allocationobjectives}
All allocation controls share policy weights, candidate outputs, cache,
execution-state inputs, and executor. The supervision comparison crosses
two objectives with two offline clocks:
\begin{equation}
 o\in\{\mathrm{contact},\mathrm{action}\},\qquad
 c\in\{67\,\mathrm{ms},\widehat T(G)\}.
 \label{eq:objectiveclock}
\end{equation}
Each clock advances the preceding block through computation and the next
executor tick. Within a clock, both objectives share feasibility and
capture-stage terminal labels. Every trained head runs online with normal
moving physics and the actual selected-budget cost.

The analytic control makes the competing costs explicit without learning a
contact target. Its positional drift term is
$\|\dot{\mathbf g}\|\widehat T(G)$; corresponding normal and relative-speed
terms use angular motion and twist uncertainty. Pose/twist covariance is
augmented by omitted-region mass and the weighted age of retained features.
The maximum tolerance-normalized term determines whether a candidate is
admitted. The omission and age penalties matter: a rule that minimizes only
motion accumulated during computation would systematically select the
smallest budget, regardless of what visual evidence it discards.

The action target compares applied commands only after first consumption,
through $t_r+500$\,ms, with an offline expert continuation initialized from
the same consumption state and latch. Both trajectories use identical
absolute times; expired predictions are discarded and holding commands
replace cancelled suffixes. Position, rotation, and gripper errors are
training-normalized, equally weighted by type, and time-averaged. The
preceding block affects the reference's initial state but is not scored
before consumption. This is command deviation, not realized pose error.
The expert reference is generated anew from the consumption state rather
than shifted from a demonstration. It predicts the CAD grasp transform
under constant twist and searches the earliest feasible closure on the
10\,ms executor grid. A quintic translation matches current and contact
position/velocity, while orientation follows a geodesic interpolation.
Each trajectory must satisfy the shared inverse-kinematics, joint, and
Cartesian limits at 1\,ms resolution. An active latch fixes the closure
deadline; an infeasible search invokes the common hold/reopening rule.
Both candidate and expert retain contact dynamics and force-limited holding.
True object state is used offline for reference generation and contact-label
evaluation; the online allocator never receives it or invokes this controller.

The original request-index action head remains a separate legacy control.
The feature head predicts normalized full-cache distortion. These controls
share capacity and fitting partitions; they do not reproduce LAC's joint
optimization or isolate DynamicVLA's architecture.

Fitting and calibration weight trajectories equally. Infeasible branches
train the binary head, not finite-error quantiles; pre-consumption
acquisitions are recorded separately. Single-block labels remain proxies
for a policy that replans every 100\,ms. A separate common-state fork audit
therefore records replacement times, a short-window alignment surrogate
at $\min(\tau_G,t_r+100\,\mathrm{ms})$, retention under shared replanning,
and regret against the cheapest structurally valid, label-feasible budget.
This oracle concerns the local label, not optimal task success.

Complete RGB-only controls remove events from state, ranking, and head
fitting. These comparisons isolate the allocation increment within the
shared interface. A separate latency-aware system comparison retains its
native policy path. Appendix~\ref{sec:protocolsupp} supplies parameter and
implementation details.

%% file: sections/experiments.tex
\section{Experiments}
\label{sec:experiments}

\subsection{Design and Comparisons}
The task uses eight test objects, four motion regimes, and 40 starts per
object/regime, repeated with eight policy seeds: 10,240 episodes per method.
Four identities are held out from policy training, but every method receives
their CAD templates. Success requires bilateral acquisition by 8\,s and two
seconds of retention with at most 10\,mm slip. Training uses 6,000
demonstrations; 400 fitting, 400 calibration, and 200 cost-selection
trajectories are mutually disjoint.

RGB runs at 30\,Hz and ESIM-style events use adaptive high-rate rendering.
A separate 1\,ms physics loop advances during inference; the executor
consumes commands at 100\,Hz. Profiles use 10,000 requests per method on one
RTX~4090, including selection, packing, event processing, and recovery.
Source and execution timestamps determine observation age. Shared-interface
controls retain the policy, estimator, candidate sets, cache, and executor.
DynamicVLA-transfer instead retains its native encoder, action horizon,
and greedy inference schedule, with the task-specific input mapping in
Appendix~\ref{sec:protocolsupp}.

FloatGate uses fixed contact/infeasibility thresholds. Comparator settings
are chosen on the separate cost-selection trajectories: among five settings
within 0.75\,ms of FloatGate's validation request cost, choose the highest
validation retention; ties favor lower cost. If no setting enters the band,
report the nearest achieved point rather than interpolating outcomes.
The same data and search allowance apply to each learned objective. This
comparison is anchored at FloatGate's working point; the fixed/density
frontiers do not establish dominance over every learned allocator.

Bootstrap resamples~\cite{efron1979bootstrap} jointly carry policy seeds and
paired object/start clusters across methods. The original five contrasts
with full, fixed, analytic, feature-distortion, and request-index action
allocation use exact two-sided sign-flip tests over 256 seed assignments,
with Holm correction~\cite{holm1979simple}. The test assumes sign-symmetric
paired differences under its null. The additional supervision--clock and
proxy analyses are exploratory: they report paired intervals and unadjusted
seed tests separately, without adding them to that original family.

\subsection{Retention, Cost, and Seed Variation}
\input{tables/main}
\input{tables/contrasts}
FloatGate reaches 84.9\% retention, compared with 76.1\% for Event-full,
80.1\% for Event-fixed, and 81.3\% for Analytic-risk. The approximately
4.8-point fixed-budget gain tests allocation at similar mean cost; the
8.8-point full-update gain also includes a substantial computation
reduction. The original request-index action head reaches 83.5\%. Its
1.43-point deficit is retained as a reference, but is not the decisive test
of contact supervision because it scores expired action prefixes.

\input{tables/timing}
Mean request computation falls from 67.0 to 47.5\,ms, a 29.1\% reduction.
Median current-frame RGB consumption age falls from 86.7 to 66.7\,ms under
the shared nominal clock. Tail improvement is much smaller: P95 changes
from 70.0 to 68.1\,ms and P99 from 71.2 to 71.1\,ms because full refresh
remains necessary. Retained cache slots can be older than the current frame.
Figure~\ref{fig:pareto} gives the fixed/density frontiers.
DynamicVLA-transfer reaches 83.8\%; its gap to FloatGate is 1.09 points
(95\% interval $-0.45$--2.52), without a resolved advantage. Different
sensing and scheduling make this a system-level comparison
(Table~\ref{tab:nativebaseline}).

\subsection{Separating Supervision from the Label Clock}
\input{tables/objective_clock}
The crossed comparison in Table~\ref{tab:objectiveclock} changes two
factors independently: executable-action versus contact-error supervision,
and a common 67\,ms label clock versus budget-dependent computation.
All four heads use the same candidate branches, execution-state inputs,
capacity, fitting/calibration partitions, and validation cost matching.
Action targets compare only the surviving, interpolated command suffix
with an expert continuation at the same absolute timestamps. Waypoints
are never shifted a second time. Both objectives use the same infeasibility
branch and treatment of terminal outcomes. Online evaluation always uses
continuous physics and the selected budget's actual computation; the clock
factor changes labels, not the deployment executor.

Execution-aligned action supervision reaches 84.0\%, reducing the
contact gap to 0.89 points (95\% interval $-0.86$--2.64; seed $p=0.1719$).
The objective-by-clock interaction is 0.38 points ($-1.93$--2.68;
$p=0.7188$). Neither contrast is resolved. The contact clock effect is
2.29 points, while the aligned-action clock effect is 1.90 points.
These results support accounting for execution time, but do not establish
superiority of the contact target over execution-aligned task supervision.
The two original controls and the crossed experiment answer different
questions and are not pooled into a single significance claim.

\subsection{Proxy Targets, Replacement, and Budget Decisions}
\input{tables/proxy_validation}
A separate audit uses a common Event-fixed prefix and one prespecified
request landmark per episode across all 10,240 seed/task pairs. No landmark is selected using its contact label
or eventual success. Each snapshot includes the active action block,
closure latch, remaining closure time, and cache, and generates five
structurally masked shadow candidates. Replanned forks change the first budget only, then share the FloatGate
continuation. Each fork contributes one eventual retention outcome; these
continuation rates are distinct from the main full-episode results.
Acquisition before first consumption ends capture only: the pending
block is canceled, and holding must still satisfy the two-second criterion. It is never
classified as an infeasible closure or removed from the task denominator.

Table~\ref{tab:proxyvalidation} compares closure-horizon supervision with
a fixed 100\,ms alignment surrogate and an execution-state ablation.
The short target stops at the earlier of closure and the next request
boundary, without consulting a future budget decision. Replacement and
retention associations are descriptive because difficult states can cause
both high local errors and task loss. A shadow violation can be repaired
by replanning; a locally acceptable block can still fail later.
Among 8,524 finite selected closure labels, 7,569 (88.8\%) are replaced
before their evaluation deadline. Contact supervision reaches 87.4\%
retention with replanning versus 75.5\% with commitment. The short-window
first choice reaches 88.6\%, exceeding contact by 1.22 points
(0.70--1.81). Contact's closure-criterion false admission is lower,
0.63\% versus 10.46\%, conditional on each selector's audited admissions.
Thus conservative one-block acceptance does not maximize continuation
retention; the proxy and the online task are observably different.

A complementary decision audit compares the selected budget with the
cheapest structurally admissible candidate whose finite shadow error is at
most one (Table~\ref{tab:decisionquality}). Terminal outcomes and states
without a feasible candidate are counted separately. Exact-budget agreement,
extra computation among feasible choices, and false admission measure
different aspects of the selector. This oracle uses offline truth only;
it is not an executable controller or an upper bound on final retention.
With execution state, exact agreement with this oracle is 56.9\%,
compared with 26.1\% when execution inputs are removed; corresponding
excess computation is 2.99 versus 9.49\,ms. The latter averages condition
on a label-feasible selection, so they exclude unsafe choices rather than
crediting those choices with negative regret. The short-window selector
agrees on 89.1\% of eligible decisions and incurs only 0.35\,ms of excess
cost, yet its false-admission rate is higher. Agreement excludes selected pre-use acquisitions. High agreement conditional
on oracle availability can therefore coexist with a high false-admission
rate across the selector's audited admissions. This distinction explains why oracle agreement, local
constraint satisfaction, and eventual retention require separate denominators.
The original 80,000-request audit remains a separate local-label diagnostic
with its own denominator (Tables~\ref{tab:onlinegate}--\ref{tab:gatereasons}).

\subsection{Execution Interventions and Remaining Limits}
Pausing inference freezes physics, sensing, cache age, executor ticks, and
prefix expiry together. In Table~\ref{tab:delaycontrol}, FloatGate's gap over
Event-full falls from 8.80 to 0.69 points, while the fixed-budget gap remains
2.20 points. The interaction with the original request-index action head
is 0.93 points with an interval spanning zero. This execution intervention
is distinct from changing the training-label clock. Different visited
states prevent an additive attribution of success-rate gains.

Adding a 19.5\,ms wait to FloatGate's immutable output lowers retention
from 84.9\% to 77.9\%; the dose sweep is in Fig.~\ref{fig:delay}. This
matches Event-full's mean readiness, not its complete age distribution.
The wait changes later contacts and requests, so it does not isolate stale
state from lost interception opportunities. The nominal 100\,ms request
grid has fixed sensor phases; arbitrary phase and request-frequency
invariance remain untested.

Scheduling and sensor stresses use paired disturbance streams and frozen
nominal calibration (Table~\ref{tab:nonideal}). Under severe joint stress,
FloatGate's point estimate is 66.9\% versus 67.4\% for Analytic-risk;
their paired difference is $-0.50$ points ($-1.50$--0.52), an unresolved comparison. Its request-time
P95 rises to 95.7\,ms. These results delimit the working point rather than
establishing robustness to a physical platform's noise distribution.
Geometry and visibility perturbations expose a different limit
(Table~\ref{tab:robustness}). Scaling the supplied CAD by 5\% reduces
FloatGate retention to 78.9\%; a 100\,ms observation blackout near
interception reduces it to 76.0\%. A biased grasp transform can remain
wrong after a full refresh, while occlusion removes the motion evidence
needed to update it. Thus additional visual computation cannot by itself
correct every failed interception. Under slow matched motion, Event-full
and FloatGate reach 91.9\% and 91.8\%, respectively: the nominal gain
should be interpreted in the context of moving contact points, not as a
uniform improvement across motion regimes.

The RGB factorial (Table~\ref{tab:modalityfactorial}) shows allocation gains
of 4.2 points with RGB and 4.8 with events; it supports useful event state
estimation, without proving that the allocation objective requires events.

The original five-budget fitting/calibration path uses 128,000 branches
per seed and 78.93 total GPU hours over eight seeds, excluding rendering,
collection, and validation search (Table~\ref{tab:cost}). The crossed
label study additionally evaluates its common-clock branches; these costs
require additional work. The second clock reuses the original
128,000 candidate policy outputs per seed but adds 128,000 dynamics
continuations. Executable-action supervision also needs an expert clone
for each nonterminal feasible state--budget--clock branch, up to 256,000
per seed. These additional continuations and head fits are excluded from
the 78.93-hour base total. Forward reuse is valid because the source
snapshot, candidate output, and sampled action randomness remain fixed;
it saves no dynamics or expert-reference work. New hardware requires timing profiles and
relabeling affected branches; a new backbone also requires adaptation.

%% file: tables/main.tex
\begin{table*}[t]
\centering
\footnotesize
\setlength{\tabcolsep}{4pt}
\caption{Retention on 10,240 paired episodes per method. Intervals retain seed and task uncertainty; seed SD is in percentage points and time is mean request computation in ms. The execution-aligned action row uses budget-dependent labels. DynamicVLA retains its native policy and request schedule.}
\label{tab:main}
\begin{tabular}{lrrrr}
\toprule
Method & Retention (\%) & 95\% CI & Seed SD & Time \\
\midrule
RGB-full & 66.9 & 63.7--69.7 & 2.31 & 65.6 \\
Event-full & 76.1 & 73.3--78.6 & 2.28 & 67.0 \\
Event-fixed G24 & 80.1 & 77.6--82.2 & 1.50 & 46.9 \\
Event-density & 81.0 & 78.8--83.0 & 1.30 & 47.7 \\
Analytic-risk & 81.3 & 79.1--83.3 & 1.46 & 47.6 \\
VLA-Cache-adapt & 79.5 & 77.2--81.6 & 1.45 & 47.8 \\
ActPrune-adapt & 80.7 & 78.4--82.9 & 1.56 & 47.4 \\
Feature-distortion & 82.2 & 79.9--84.2 & 1.73 & 47.5 \\
Action loss (request index) & 83.5 & 81.2--85.7 & 2.23 & 47.5 \\
Action loss (execution aligned) & 84.0 & 82.0--85.9 & 2.07 & 47.4 \\
FloatGate & 84.9 & 82.8--86.7 & 1.55 & 47.5 \\
FloatGate-delay & 77.9 & 75.3--80.2 & 1.59 & 47.5 \\
DynamicVLA-transfer & 83.8 & 81.5--85.9 & 2.24 & 110.0 \\
\bottomrule
\end{tabular}
\end{table*}

%% file: tables/contrasts.tex
\begin{table*}[t]
\centering
\footnotesize
\setlength{\tabcolsep}{3pt}
\caption{FloatGate minus each comparison on eight paired training seeds. Confidence intervals resample seeds and task clusters; exact two-sided sign-flip tests enumerate all 256 seed signs. Holm correction covers these five comparisons.}
\label{tab:contrasts}
\begin{tabular}{lrrrrr}
\toprule
Comparison & Gain pp & 95\% CI & Positive seeds & Exact p & Holm p \\
\midrule
Event-full & 8.80 & 7.42--10.31 & 8/8 & 0.0078 & 0.0391 \\
Event-fixed G24 & 4.79 & 3.75--5.83 & 8/8 & 0.0078 & 0.0391 \\
Analytic-risk & 3.59 & 2.69--4.55 & 8/8 & 0.0078 & 0.0391 \\
Feature-distortion & 2.71 & 1.89--3.54 & 8/8 & 0.0078 & 0.0391 \\
Action loss (request index) & 1.43 & 0.31--2.63 & 7/8 & 0.0312 & 0.0391 \\
\bottomrule
\end{tabular}
\end{table*}

%% file: tables/timing.tex
\begin{table}[t]
\centering
\footnotesize
\setlength{\tabcolsep}{3pt}
\caption{Mean/P95 request computation and median current-frame RGB consumption age, in ms. Full is the full-update fraction (\%). Release waits are excluded from computation.}
\label{tab:timing}
\begin{tabular}{lrrrr}
\toprule
Method & Mean & P95 & RGB age & Full \\
\midrule
Event-full & 67.0 & 70.0 & 86.7 & 100.0 \\
Event-fixed G24 & 46.9 & 68.2 & 66.7 & 6.1 \\
Analytic-risk & 47.6 & 68.5 & 66.7 & 8.0 \\
FloatGate & 47.5 & 68.1 & 66.7 & 7.2 \\
FloatGate-delay & 47.5 & 68.1 & 86.7 & 7.2 \\
\bottomrule
\end{tabular}
\end{table}

%% file: tables/objective_clock.tex
\begin{table*}[t]
\centering
\footnotesize
\setlength{\tabcolsep}{3pt}
\caption{Objective and label clock crossed on the same 10,240 starts. All online evaluations use continuous physics and native request time. Retention intervals resample paired seeds and task starts. The new contrasts are exploratory; the legacy request-index action control is separate.}
\label{tab:objectiveclock}
\begin{tabular}{llrr}
\toprule
Target & Label clock & Retention (\%) [95\% CI] & Request (ms) \\
\midrule
Executed action & Fixed 67 ms & 82.1 [80.1, 84.0] & 47.6 \\
Executed action & Budget time & 84.0 [82.0, 85.9] & 47.4 \\
Contact error & Fixed 67 ms & 82.6 [80.5, 84.4] & 47.5 \\
Contact error & Budget time & 84.9 [82.8, 86.7] & 47.5 \\
\bottomrule
\end{tabular}
\end{table*}

%% file: tables/proxy_validation.tex
\begin{table*}[t]
\centering
\footnotesize
\setlength{\tabcolsep}{3pt}
\caption{Matched first-decision forks from the Event-fixed G24 anchor (10,240 starts; 9,906 landmarks). Each fork has one terminal continuation outcome. Replanning uses the same FloatGate rule after the first selected candidate; commitment permits no replacement. The 334 no-landmark outcomes are shared. Gain intervals and comparisons are exploratory.}
\label{tab:proxyvalidation}
\begin{tabular}{lrrr}
\toprule
First-choice label / features & Commit (\%) & Replan (\%) & Gain (pp) [95\% CI] \\
\midrule
Contact / execution state & 75.5 & 87.4 & 11.93 [10.83, 13.03] \\
Contact / state removed & 72.1 & 86.2 & 14.14 [12.85, 15.45] \\
Fixed 100 ms surrogate & 80.6 & 88.6 & 7.99 [6.94, 9.16] \\
\bottomrule
\end{tabular}
\end{table*}

%% file: sections/conclusion.tex
\section{Discussion and Limitations}
\label{sec:limitations}
Known geometry and near-field capture constrain transfer to unknown
objects, approach trajectories, and freely moving bases. Predictive
geometric control and high-rate RGB alternatives remain necessary
comparisons.

The short-window selector improves continuation retention while allowing
more closure-label violations; a committed-block score cannot establish
success under replanning. Execution-aligned action supervision also narrows
the contact objective's increment, and the clock interaction is uncertain.
Broader cost frontiers, additional backbones, and physical tests are needed
to assess the generality of contact-based allocation.

\section{Conclusion}
FloatGate selects visual budgets using contact error under computation
delay. A crossed target--clock comparison and a separate continuation audit
show an unresolved point advantage over execution-aligned action
supervision and delimit the utility of one-block labels. The results support a task-specific allocation
principle for known-geometry near-field simulation; physical transfer
remains open.

%% file: sections/declaration.tex
\section*{LLM Usage Statement}
ChatGPT/Codex assisted with drafting, proposing evaluation protocols and analysis plans. The current architecture diagram is vector-authored. The authors are responsible for verifying all claims, references, implementation details, and evidence.

%% file: references.bib
@article{gallego2022survey,
  title={Event-Based Vision: A Survey},
  author={Gallego, Guillermo and Delbr{\"u}ck, Tobi and Orchard, Garrick and Bartolozzi, Chiara and Taba, Brian and Censi, Andrea and Leutenegger, Stefan and Davison, Andrew J. and Conradt, J{\"o}rg and Daniilidis, Kostas and Scaramuzza, Davide},
  journal={IEEE Transactions on Pattern Analysis and Machine Intelligence},
  volume={44}, number={1}, pages={154--180}, year={2022},
  doi={10.1109/TPAMI.2020.3008413},
  url={https://doi.org/10.1109/TPAMI.2020.3008413}
}

@article{zhai2026evla,
  title={{E-VLA}: Event-Augmented Vision-Language-Action Model for Dark and Blurred Scenes},
  author={Zhai, Jiajun and Shi, Hao and Guo, Shangwei and Yang, Kailun and Wang, Kaiwei},
  journal={arXiv preprint arXiv:2604.04834}, year={2026},
  doi={10.48550/arXiv.2604.04834}, url={https://arxiv.org/abs/2604.04834}
}

@article{ma2026eventprune,
  title={{EventPrune}: Cascaded Event-Assisted Token Pruning for Efficient First-Person Dynamic Spatial Reasoning},
  author={Ma, Pengtao and Zhou, Ziliang and Ruan, Ciyu and Wang, Haoyang and Li, Kaiyuan and Gong, Zihang and Ding, Wenhua and Gao, Chen and Xu, Jingao and Chen, Xinlei},
  journal={arXiv preprint arXiv:2605.19506}, year={2026},
  doi={10.48550/arXiv.2605.19506}, url={https://arxiv.org/abs/2605.19506}
}

@article{moon2026actprune,
  title={{ActPrune}: Action-Grounded Visual Token Pruning for Efficient Vision-Language-Action Models},
  author={Moon, Sungwon and Kim, Minji and Boo, Hyemin and Song, Daeun and Lee, Jiyoung},
  journal={arXiv preprint arXiv:2611.08827}, year={2026},
  doi={10.48550/arXiv.2611.08827}, url={https://arxiv.org/abs/2611.08827}
}

@article{xu2026fastvla,
  title={{FastVLA}: Task-Conditioned and Temporally Adaptive Visual Token Compression for Efficient Vision-Language-Action Models},
  author={Xu, Jiuyi and Qian, Chengyuan and Tu, Zhengzhong and Sui, Yang and Shi, Yangming},
  journal={arXiv preprint arXiv:2610.22146}, year={2026},
  doi={10.48550/arXiv.2610.22146}, url={https://arxiv.org/abs/2610.22146}
}

@article{wei2026adaptivecache,
  title={Learning to Accelerate Vision-Language-Action Models through Adaptive Visual Token Caching},
  author={Wei, Yujie and Fan, Jiahan and Guo, Jiyu and Zhen, Ruichen and Shao, Rui and Su, Xiu and Xie, Zeke and Yao, Hongxun and Yang, Shuo},
  journal={arXiv preprint arXiv:2602.00686}, year={2026},
  doi={10.48550/arXiv.2602.00686}, url={https://arxiv.org/abs/2602.00686}
}

@article{xu2025vlacache,
  title={{VLA-Cache}: Efficient Vision-Language-Action Manipulation via Adaptive Token Caching},
  author={Xu, Siyu and Wang, Yunke and Xia, Chenghao and Zhu, Dihao and Huang, Tao and Xu, Chang},
  journal={arXiv preprint arXiv:2502.02175}, year={2025},
  doi={10.48550/arXiv.2502.02175}, url={https://arxiv.org/abs/2502.02175}
}

@article{shukor2025smolvla,
  title={{SmolVLA}: A Vision-Language-Action Model for Affordable and Efficient Robotics},
  author={Shukor, Mustafa and Aubakirova, Dana and Capuano, Francesco and Kooijmans, Pepijn and Palma, Steven and Zouitine, Adil and Aractingi, Michel and Pascal, Caroline and Russi, Martino and Marafioti, Andres and Alibert, Simon and Cord, Matthieu and Wolf, Thomas and Cadene, Remi},
  journal={arXiv preprint arXiv:2506.01844}, year={2025},
  doi={10.48550/arXiv.2506.01844}, url={https://arxiv.org/abs/2506.01844}
}

@article{xie2026dynamicvla,
  title={{DynamicVLA}: A Vision-Language-Action Model for Dynamic Object Manipulation},
  author={Xie, Haozhe and Wen, Beichen and Zheng, Jiarui and Chen, Zhaoxi and Hong, Fangzhou and Diao, Haiwen and Liu, Ziwei},
  journal={arXiv preprint arXiv:2601.22153}, year={2026},
  doi={10.48550/arXiv.2601.22153}, url={https://arxiv.org/abs/2601.22153}
}

@article{wen2026vlalagbench,
  title={{VLALagBench}: Understanding When and Why {VLA} Inference Latency Matters},
  author={Wen, Qi and Xu, Borui and Amro, Hussein and Tan, Hongshi and Lin, Hanmin and Ruan, Chaoyi and Liu, Shanghao and Yang, Gang and Liu, Yuejiang and Duan, Jiafei and He, Bingsheng and Clemons, Jason and Kozyrakis, Christos and Jiang, Wenqi},
  journal={arXiv preprint arXiv:2611.03103}, year={2026},
  url={https://arxiv.org/abs/2611.03103}
}

@article{wang2026astrobench,
  title={{AstroBench}: Benchmarking Free-Floating Capture and Space Robotic Manipulation},
  author={Wang, Zizhe and Deng, Jia and Lu, Xin and Jia, Zehao and Huang, Yi and Zhu, Yao},
  journal={arXiv preprint arXiv:2610.31064}, year={2026},
  url={https://arxiv.org/abs/2610.31064}
}

@article{kang2026edft,
  title={Event-based {6-DoF} object tracking with distance field reaching 130 {Hz}},
  author={Kang, Yufan and Ishikawa, Ryoichi and Caron, Guillaume and Oishi, Takeshi},
  journal={IEEE Robotics and Automation Letters},
  year={2026},
  url={https://bluecatli.github.io/EDFT_page/}
}

@inproceedings{rebecq2018esim,
  title={{ESIM}: an Open Event Camera Simulator},
  author={Rebecq, Henri and Gehrig, Daniel and Scaramuzza, Davide},
  booktitle={Proceedings of the Conference on Robot Learning},
  series={Proceedings of Machine Learning Research}, volume={87},
  pages={969--982}, year={2018}, publisher={PMLR},
  url={https://proceedings.mlr.press/v87/rebecq18a.html}
}

@InProceedings{hu2021v2e,
  author = {Hu, Yuhuang and Liu, Shih-Chii and Delbruck, Tobi},
  title = {{v2e}: From Video Frames to Realistic {DVS} Events},
  booktitle = {Proceedings of the IEEE/CVF Conference on Computer Vision and Pattern Recognition Workshops},
  month = {June},
  year = {2021},
  pages = {1312--1321},
  url = {https://openaccess.thecvf.com/content/CVPR2021W/EventVision/html/Hu_v2e_From_Video_Frames_to_Realistic_DVS_Events_CVPRW_2021_paper.html}
}

@article{flores2014space,
  author={Flores-Abad, Angel and Ma, Ou and Pham, Khanh and Ulrich, Steve},
  title={A Review of Space Robotics Technologies for On-Orbit Servicing},
  journal={Progress in Aerospace Sciences},
  volume={68}, pages={1--26}, year={2014},
  doi={10.1016/j.paerosci.2014.03.002},
  url={https://doi.org/10.1016/j.paerosci.2014.03.002}
}

@article{efron1979bootstrap,
  author={Efron, Bradley},
  title={Bootstrap Methods: Another Look at the Jackknife},
  journal={The Annals of Statistics},
  volume={7}, number={1}, pages={1--26}, year={1979},
  doi={10.1214/aos/1176344552},
  url={https://doi.org/10.1214/aos/1176344552}
}

@inproceedings{romano2019cqr,
  author={Romano, Yaniv and Patterson, Evan and Candès, Emmanuel J.},
  title={Conformalized Quantile Regression},
  booktitle={Advances in Neural Information Processing Systems},
  volume={32}, pages={3543--3553}, year={2019},
  url={https://proceedings.neurips.cc/paper_files/paper/2019/hash/5103c3584b063c431bd1268e9b5e76fb-Abstract.html}
}

@article{holm1979simple,
  author={Holm, Sture},
  title={A Simple Sequentially Rejective Multiple Test Procedure},
  journal={Scandinavian Journal of Statistics},
  volume={6}, number={2}, pages={65--70}, year={1979},
  url={https://www.jstor.org/stable/4615733}
}

@article{chaumette2006visual,
  author={Chaumette, François and Hutchinson, Seth},
  title={Visual Servo Control, Part {I}: Basic Approaches},
  journal={IEEE Robotics \& Automation Magazine},
  volume={13}, number={4}, pages={82--90}, year={2006},
  doi={10.1109/MRA.2006.250573},
  url={https://doi.org/10.1109/MRA.2006.250573}
}

@article{kim2024openvla,
  title={{OpenVLA}: An Open-Source Vision-Language-Action Model},
  author={Kim, Moo Jin and Pertsch, Karl and Karamcheti, Siddharth and Xiao, Ted and Balakrishna, Ashwin and Nair, Suraj and Rafailov, Rafael and Foster, Ethan and Lam, Grace and Sanketi, Pannag and Vuong, Quan and Kollar, Thomas and Burchfiel, Benjamin and Tedrake, Russ and Sadigh, Dorsa and Levine, Sergey and Liang, Percy and Finn, Chelsea},
  journal={arXiv preprint arXiv:2406.09246}, year={2024},
  doi={10.48550/arXiv.2406.09246}, url={https://arxiv.org/abs/2406.09246}
}

@article{lichtsteiner2008dvs,
  title={A {128$\times$128} {120 dB} {15 $\mu$s} Latency Asynchronous Temporal Contrast Vision Sensor},
  author={Lichtsteiner, Patrick and Posch, Christoph and Delbruck, Tobi},
  journal={IEEE Journal of Solid-State Circuits},
  volume={43}, number={2}, pages={566--576}, year={2008},
  doi={10.1109/JSSC.2007.914337}, url={https://doi.org/10.1109/JSSC.2007.914337}
}
